\documentclass{article}

\usepackage[utf8]{inputenc}
\usepackage[english]{babel}
\usepackage[numbers]{natbib}
\usepackage[colorlinks=true, linkcolor=blue, citecolor=blue, urlcolor=blue]{hyperref}

\usepackage{listings}
\usepackage{amsmath, amssymb, algorithm, algcompatible}
\usepackage{graphicx}
\usepackage{multirow}
\usepackage{enumitem}
\usepackage{multicol}
\usepackage{enumitem}
\usepackage{float}
\usepackage{amsmath,amssymb,amsfonts}
\usepackage{amsthm}
\usepackage{mathrsfs}
\usepackage[title]{appendix}
\usepackage{xcolor}
\usepackage{textcomp}
\usepackage{manyfoot}
\usepackage{booktabs}
\usepackage{caption}
\usepackage{algorithm}
\usepackage{algorithmicx}
\usepackage{algpseudocode}
\usepackage{listings}
\usepackage{hyperref}
\usepackage{amssymb}
\usepackage{booktabs}
\usepackage{array}
\usepackage{xcolor}
\usepackage{upquote} 
\usepackage{subcaption}
\usepackage{float}
\usepackage{tabularx}
\usepackage{lipsum}
\usepackage{rotating}
\usepackage{paralist}
\usepackage{appendix}
\usepackage{dirtytalk}
\usepackage{csquotes}
\usepackage{array}
\usepackage{siunitx}
\begin{document}
\title{Analyzing the Impact of Fake News on the Anticipated Outcome of the 2024 Election Ahead of Time}

\author{Shaina Raza\thanks{Vector Institute for Artificial Intelligence, Toronto, ON, Canada. Email: shaina.raza@vectorinstitute.ai, ORCID: 0000-0003-1061-5845} \and Mizanur Rahman\thanks{Royal Bank of Canada, Toronto, ON, Canada. Email: mizanur.york@gmail.com} \and Shardul Ghuge\thanks{Vector Institute for Artificial Intelligence, Toronto, ON, Canada. Email: shardulghuge07@gmail.com, ORCID: 0000-0001-0728-0733}}

\maketitle

\begin{abstract}
Despite increasing awareness and research around fake news, there is still a significant need for datasets that specifically target racial slurs and biases within North American political speeches. This is particulary important in the context of upcoming North American elections. This study introduces a comprehensive dataset that illuminates these critical aspects of misinformation. To develop this fake news dataset, we scraped and built a corpus of 40,000 news articles about political discourses in North America. A portion of this dataset (4000) was then carefully annotated, using a blend of advanced language models and human verification methods. We have made both these datasets openly available to the research community and have conducted benchmarking on the annotated data to demonstrate its utility. We release the best-performing language model along with data. We encourage researchers and developers to make use of this dataset and contribute to this ongoing initiative.

\end{abstract}



\section{Introduction}

The term ``fake news" denotes misinformation or disinformation, which often disguises as legitimate news\cite{lazer2018science,raza2022fake}. The fake news content often spread swiftly through social media and other online platforms and leads to negative consequences in the real world. The implications of fake news are far-reaching, impacting public opinion, influencing policy decisions, and potentially altering the course of democratic processes \cite{raza2022news}.

Fake news has been a factor in many events such as elections, where it has been known to influence voter perceptions and decisions \cite{raza2021automatic}. A prominent example is the 2016 United States presidential election, which saw an influx of false stories potentially affecting the electorate's views \cite{allcott2017social}. Similar instances have been observed in US Elections 2020 \cite{pennycook2021examining}, where fabricated news has been employed to sway voter opinion and impacted the outcomes.

Artificial Intelligence (AI) and Natural Language Processing (NLP) are crucial in countering fake news by analyzing and detecting misinformation \cite{qi2019exploiting} in bulk of news. The machine learning algorithms (ML) can easily sift through vast data to identify fake news patterns \cite{raza2021automatic}. NLP can also analyzes text for authenticity, bias, and source credibility \cite{raza2022fake}.  The incorporation of these advanced methods into news media portals has the potential to cultivate a truth-centric information ecosystem \cite{muhammed_t_disaster_2022-3}.

In preparation for the \textit{upcoming 2024 elections in North America}, including the US presidential and Canadian federal elections, it is important to construct a targeted dataset specifically focused on fake news. This dataset will fill a significant void by capturing the evolving tactics of contemporary misinformation and digital media trends, which have developed since previous elections. Our goal with this work is to safeguard democratic integrity and promotes well-informed public discourse.

\paragraph{Previous Works}
``Fake news'' encompasses various forms of misleading information, including misinformation, clickbait, and disinformation \citep{allcott2017social, raza2022fake}. Advances in ML and DL have led to significant progress in fake news detection, with a range of datasets \citep{yetim2021fakenewsdetection, dulizia2021repository, hossain2021banfakenews, nic2021covid19fakenews, shu2017fake, kaggleafnd2021, kagglesarcasmdetection2021} developed for this purpose. These datasets, however, face challenges such as `concept drift' or `data drift', where the nature of misinformation changes over time, especially during events like elections. 

Most existing datasets focus on sentence-level fake news detection \citep{babu2023performance, khan2021benchmark, kumar2020fake, oshikawa2018survey, villela2023fake}. However, what is missing is identifying specific linguistic indicators of fake news. Also, most of these datasets are typically annotated by either crowdsourced workers (non-experts) or domain experts. Recent advancements in large language models (LLMs) \citep{chang2023survey, zhao2023survey} have enhanced data annotation accuracy\cite{gilardi2023chatgpt}, where the LLM annotate data and experts can review that. Our study is inspired by the effectiveness and accuracy of a hybrid approach combining LLMs with expert input, as evidenced in the recent works \cite{kuzman2023chatgpt,goel2023llms}.

\paragraph{Our Contributions}
In this work, we present a fake news dataset centered around political discourse. Our specific contributions are: 
\\
\newline
 (1)  We have created a comprehensive dataset focused on racial slurs and biases in North American political speeches. This dataset was compiled from May to October 2023, using a systematic approach to categorize and extract news articles. These articles, selected using keywords such as 'race, politics, religion, and gender,' emphasize societal issues. The dataset not only provides text from these articles but also identifies instances of fake news through specific words and phrases and aspects, illustrating the spread of disinformation and misinformation.\\
\newline
 (2) The dataset encompasses approximately 40,000 news articles, with 4,000 of them being thoroughly annotated. This annotation was accomplished using OpenAI's API \cite{openai}, with added human verification to ensure data quality and reliability.\\
\newline
  (3) These datasets are made publicly available, inviting both use and contributions from the research community. They can be accessed at:
    
    \textbf{Fake News Elections 2024 Dataset}: \href{https://huggingface.co/datasets/newsmediabias/fake\_news\_elections2024}{fake\_news\_2024}.
    
     \textbf{Annotated Fake News Elections Dataset}: \href{https://huggingface.co/datasets/newsmediabias/fake\_news\_elections\_labelled\_data}{annotated\_fake\_news}

\section{Dataset Construction}

\paragraph{Data Extraction and Keywords:} In developing our dataset, we aimed to capture a wide array of news articles addressing racial slurs and biases in North American political speeches. Utilizing the `feedparser` \cite{feedparser_pypi} and `newspaper.Article` API \cite{newspaper_docs}, we extracted essential information from each article, including the text (snippets), publication date, news outlet, and source URL.
 
\paragraph{Keywords:} Our dataset employs a strategically diverse set of keywords, covering a broad spectrum of themes. This includes political and economic terms (like immigration reform, tax legislation), social and cultural dimensions (such as gender equality), environmental and scientific topics (renewable energy, climate change), global affairs (Brexit, United Nations), historical references (the Civil Rights Movement, the Cold War), and emerging social issues (cryptocurrency regulation, data privacy). 

\paragraph{Overview of Data Collection and Annotation Process:} The overview of data collection and annotation process is given in Table \ref{tab:data_collection_annotation}
\begin{table}
\centering
\small
\begin{tabular}{|l|p{7cm}|}
\hline
\textbf{Aspect}                  & \textbf{Description} \\ \hline
Time Frame                       & Data collection covered a six-month period, from May to October 2023, to capture temporal variations.\\ \hline
Inclusion Criteria               & Articles with relevant keywords, published within the study period, and fully accessible. \\ \hline
Exclusion Criteria               & Articles with incomplete text, non-English content (focusing on North American discourse), and redundant articles. \\ \hline
Total Articles Scraped           & 40,000 articles collected. \\ \hline
Articles Chosen for Annotation   & 4,000 articles chosen for in-depth annotation, representing a wide topic range. \\ \hline
Annotation Method                & Combination of OpenAI's GPT-Turbo-3.5 API for initial AI-driven annotations and expert human verification for refinement.\\ \hline
\end{tabular}
\caption{Overview of Data Collection and Annotation Process}
\label{tab:data_collection_annotation}
\end{table}
\paragraph{Dataset Schema:}
The dataset has following fields, shown in Table \ref{table:main_dataset_fields}:
\begin{table}[H]

\centering
\begin{tabular}{|l|p{8cm}|}
\hline
\textbf{Field} & \textbf{Description} \\
\hline
\textit{text} & Main content of the news or article, conveying the core story. \\
\hline
\textit{source} & URL or original source of the news, crucial for source validation. \\
\hline
\textit{date\_published} & Publication date of the news article, offering temporal context. \\
\hline
\textit{keyword\_category} & Categorization based on associated keywords for thematic analysis. \\
\hline
\textit{outlet} & Name of the publishing news outlet, indicating potential biases. \\
\hline
\end{tabular}
\caption{Main Dataset Fields}
\label{table:main_dataset_fields}
\end{table}

The labelled dataset comprises the following fields, shown in Table \ref{tab:data_collection_annotation}:
\begin{table}[h]
\centering
\begin{tabular}{|l|p{8cm}|}
\hline
\textbf{Field} & \textbf{Description} \\
\hline
\textit{dimension} & Indicates the dimension of fake news (e.g., political, social). \\
\hline
\textit{aspect} & Captures specific news aspects (may contain NaN values for missing data). \\
\hline
\textit{fake\_phrases} & Identifies fake phrases in the text for textual analysis. \\
\hline
\textit{debias\_text} & Modified text with removed/altered fake phrases for debiased view. \\
\hline
\textit{label\_fake} & Binary indicator for news authenticity (fake or real). \\
\hline
\end{tabular}
\caption{Annotated Dataset Fields}
\label{table:annotated_dataset_fields}
\end{table}

\section{Data Analysis}

\begin{figure}[htbp]
\centering
\begin{subfigure}[t]{0.49\textwidth}
    \includegraphics[width=\textwidth]{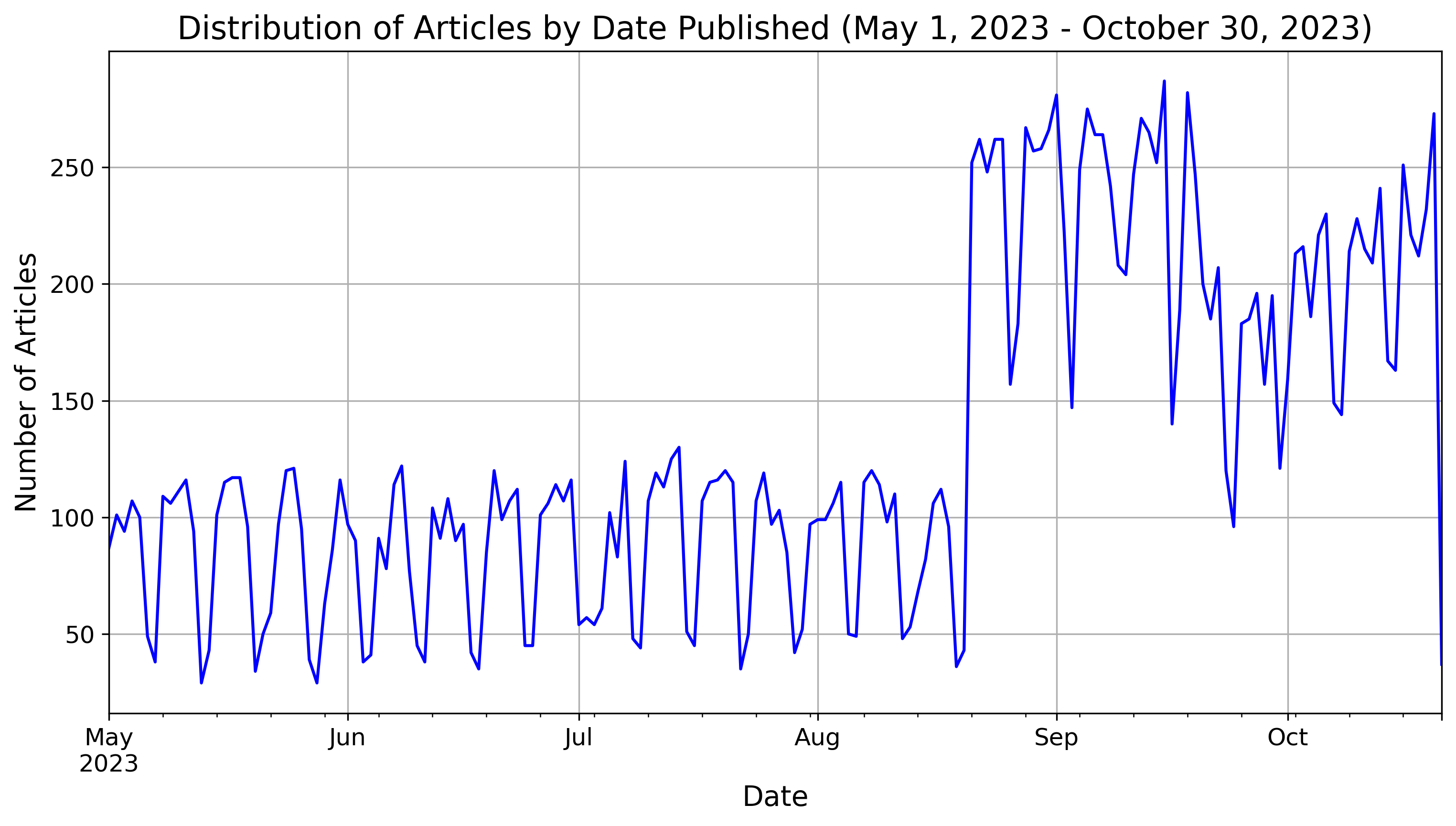}
    \caption{Distribution of Articles from May to October 2023}
    \label{fig:may_to_nov_articles}
\end{subfigure}
\hfill
\begin{subfigure}[t]{0.49\textwidth}
    \includegraphics[width=\textwidth]{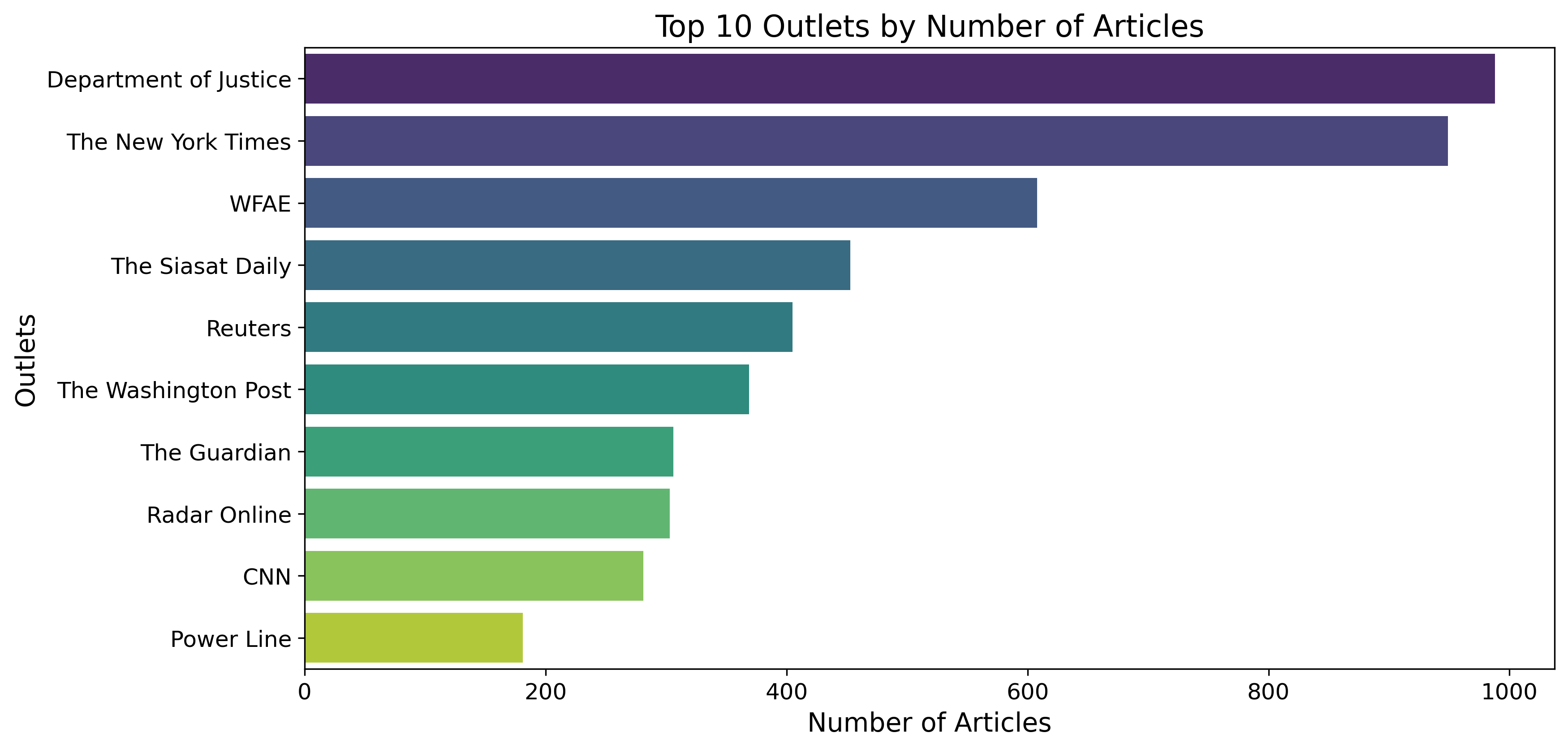}
    \caption{Top 10 News Outlets by Article Count}
    \label{fig:top_outlets}
\end{subfigure}
\\ 
\begin{subfigure}[t]{0.49\textwidth}
    \includegraphics[width=\textwidth]{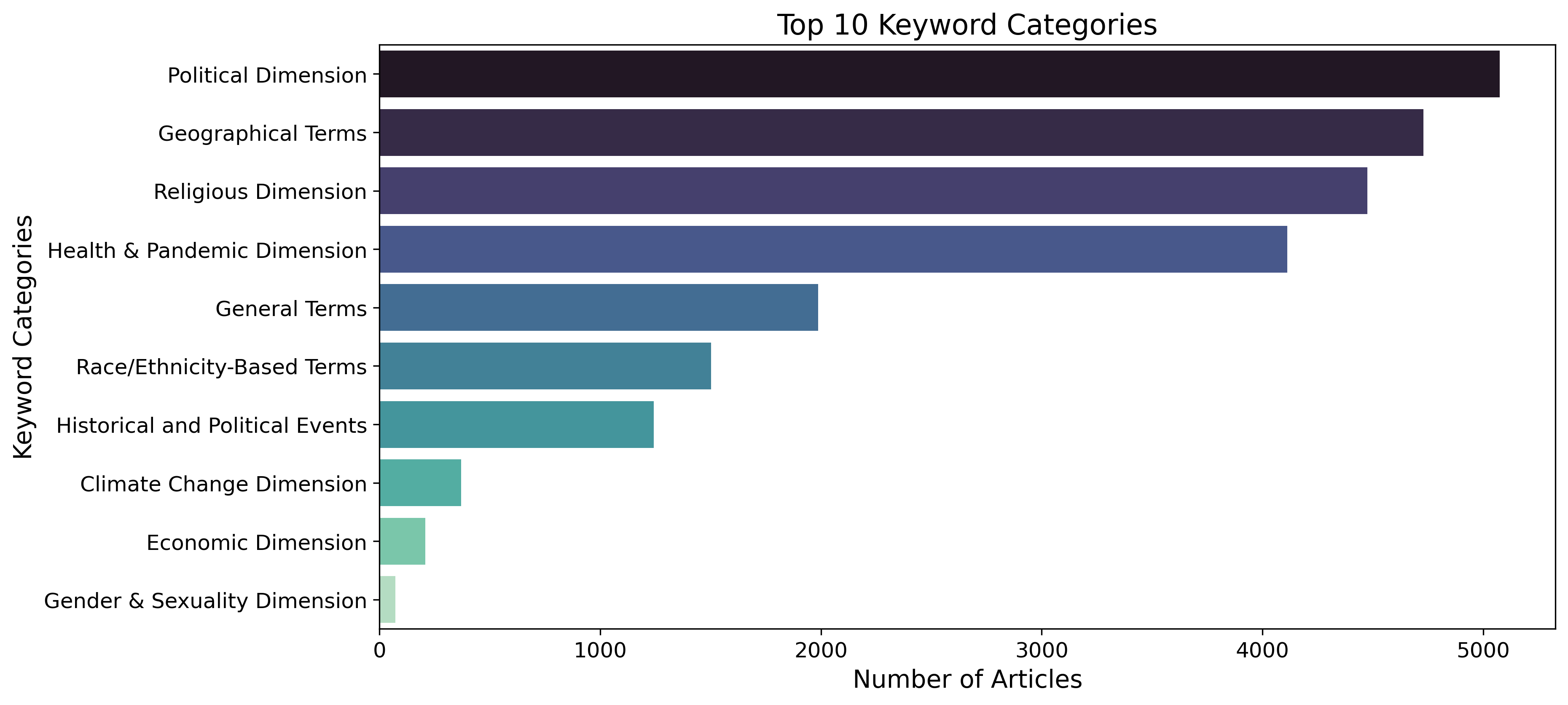}
    \caption{Top 10 Keyword Categories in News Articles}
    \label{fig:keyword_categories}
\end{subfigure}
\hfill
\begin{subfigure}[t]{0.49\textwidth}
    \includegraphics[width=\textwidth]{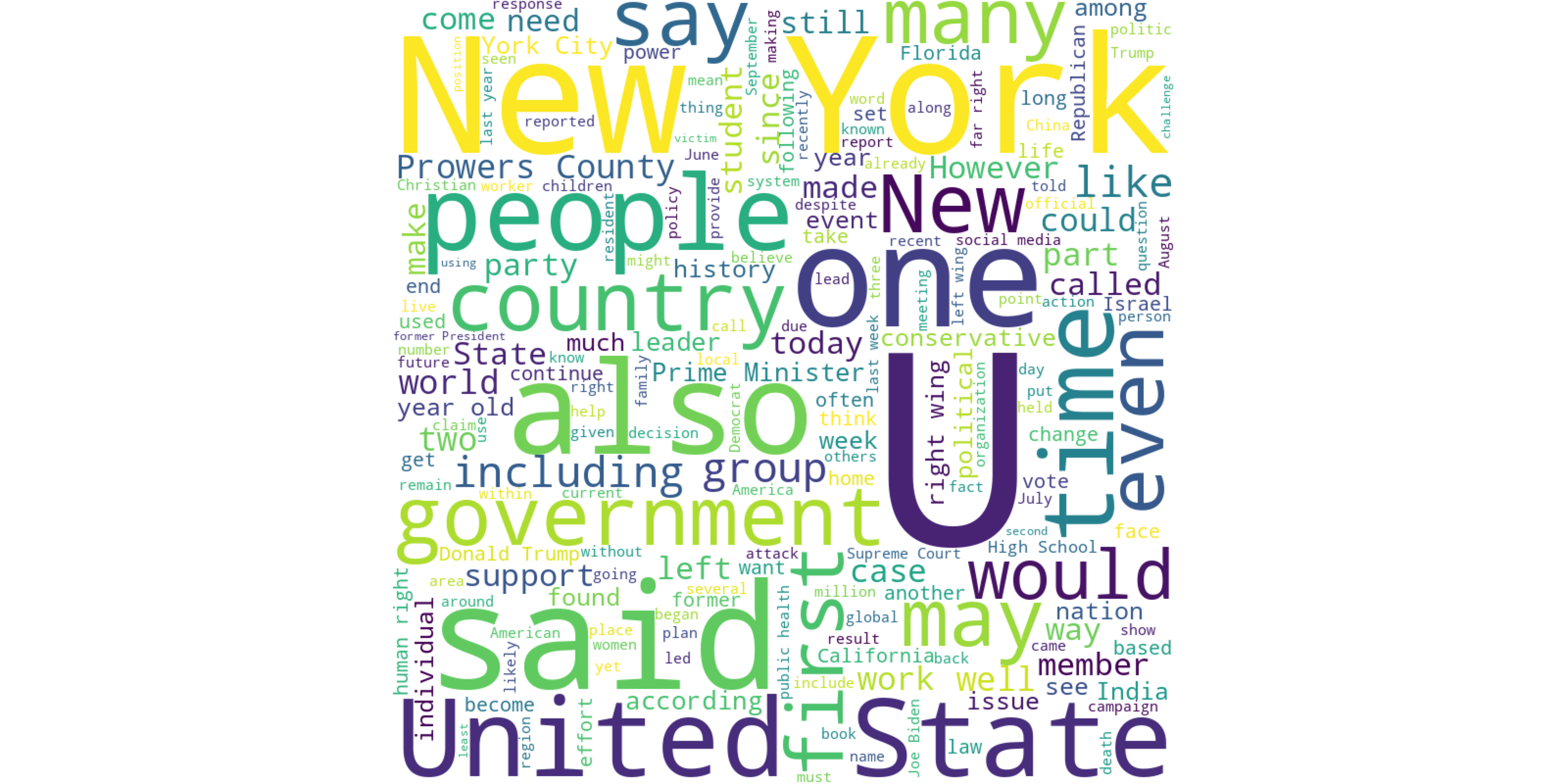}
    \caption{Word Cloud of the Most Frequent Terms in Articles}
    \label{fig:word_cloud}
\end{subfigure}
\\ 
\begin{subfigure}[t]{0.49\textwidth}
    \includegraphics[width=\textwidth]{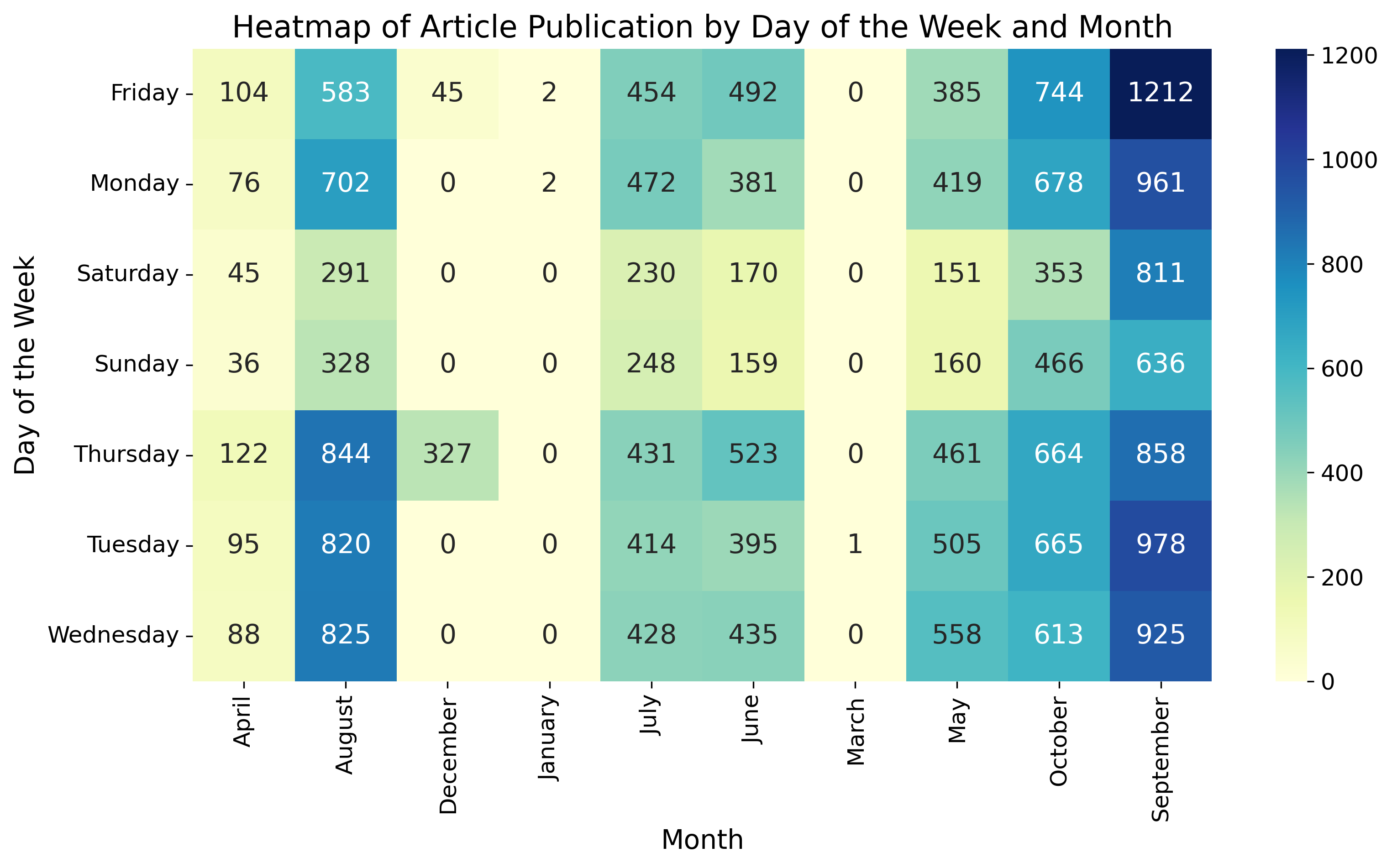}
    \caption{Heatmap of Article Publication by Day of the Week and Month}
    \label{fig:heatmap_publication}
\end{subfigure}
\hfill
\begin{subfigure}[t]{0.49\textwidth}
    \includegraphics[width=\textwidth]{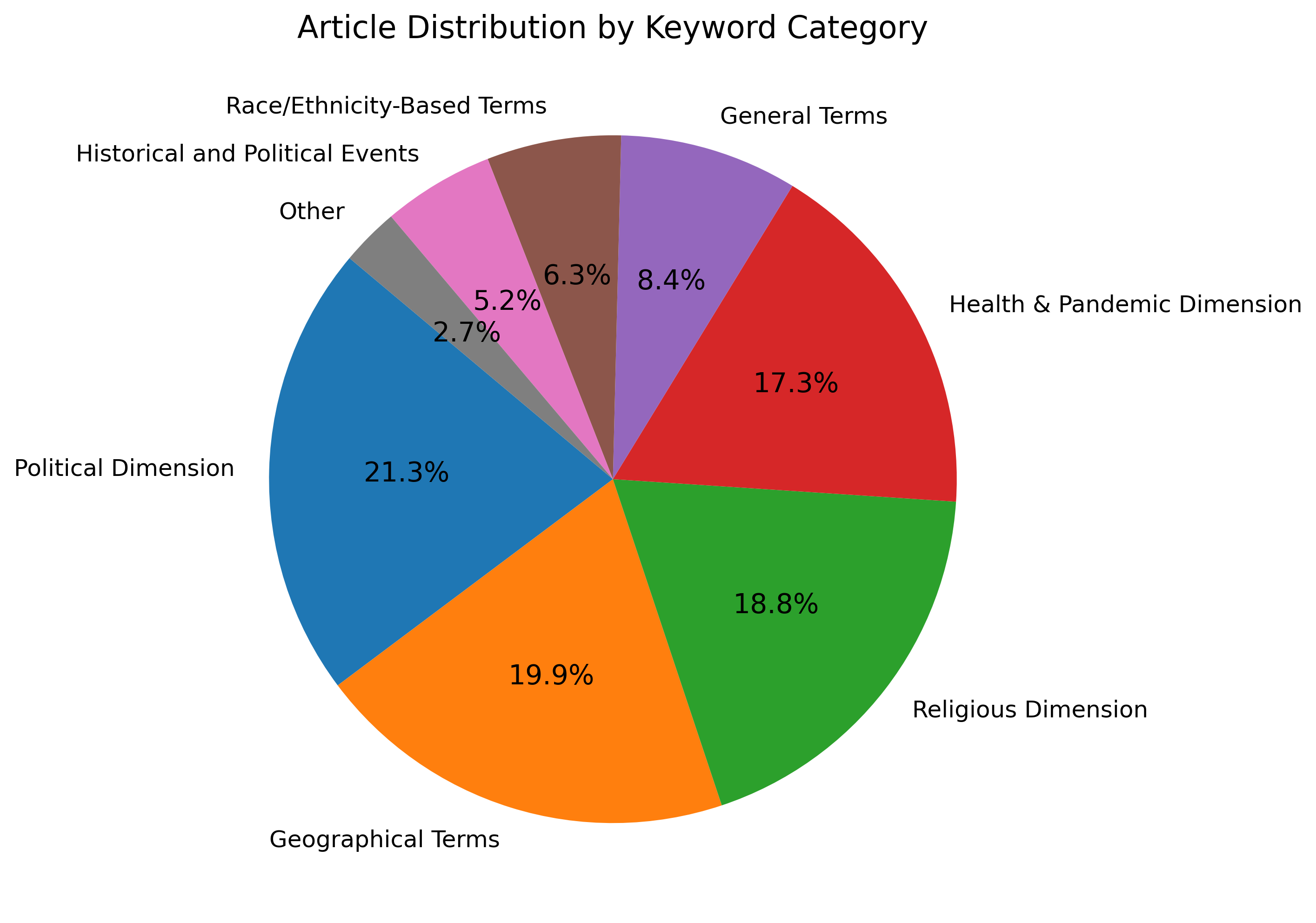}
    \caption{Pie Chart of Article Distribution by Keyword Category}
    \label{fig:pie_chart}
\end{subfigure}
\caption{Data Analysis Visualizations}
\label{fig:data_analysis_composite}
\end{figure}

We conducted a comprehensive data analysis of collected news articles, as depicted in Figure \ref{fig:data_analysis_composite}. We observe the distribution of articles over time in Figure \ref{fig:may_to_nov_articles} that reveals a notable increase in articles related to the elections during the later months, particularly in September and October. This trend suggests heightened media attention as the election date approaches. 

Upon examining the prominence of news outlets in terms of article count, as shown in Figure \ref{fig:top_outlets}, we identified the Department of Justice and The New York Times among the most prolific publishers. This comparison sheds light on which outlets were most actively covering the election. The analysis of keyword categories, detailed in Figure \ref{fig:keyword_categories}, highlights the focus on political and geographic keywords within the articles. This aspect underscores the thematic orientation of the news content during the election period.

The  word cloud in Figure \ref{fig:word_cloud} visually represents the most frequent terms in the dataset, with words like `people',  `United States', and  `New York politics' being particularly prominent. This visualization shows insights into the recurring themes and focal points in the news coverage. The heatmap (Figure \ref{fig:heatmap_publication}) reveals a higher concentration of article publications on weekdays, with particularly elevated activity on Tuesdays and Thursdays. The pie chart  (Figure \ref{fig:pie_chart}) demonstrates that political news constitutes the largest segment of coverage. it shows for over 40\% of all articles, highlighting the media's strong focus on political topics during this period.

\begin{figure}[ht]
    \centering
    \begin{minipage}{0.5\textwidth}
        \includegraphics[width=\linewidth]{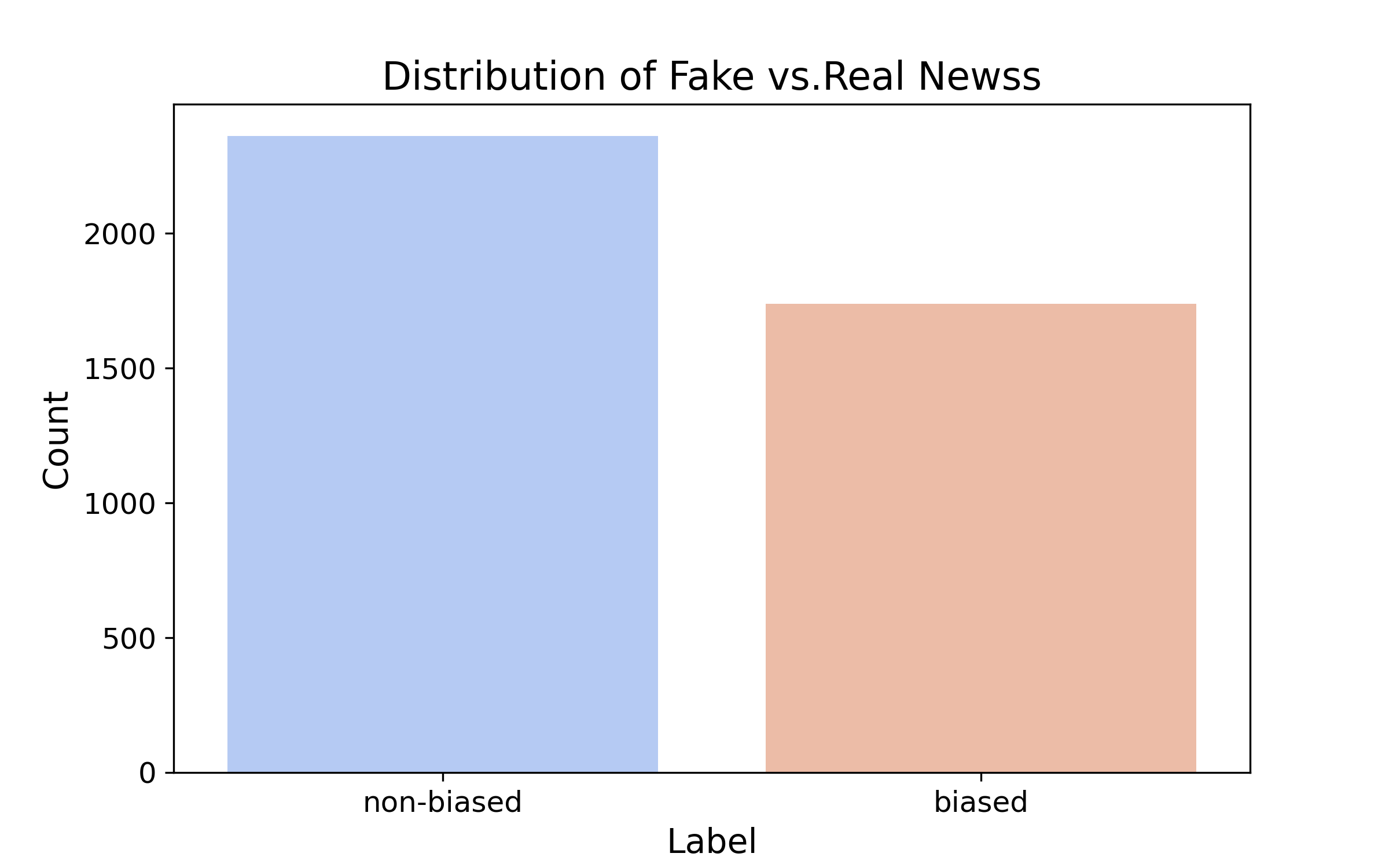}
        \captionsetup{width=1\linewidth}
        \caption{Fake vs. Real Labels}
        \label{fig:biased_non_biased}
    \end{minipage}\hfill
    \begin{minipage}{0.5\textwidth}
        \includegraphics[width=\linewidth]{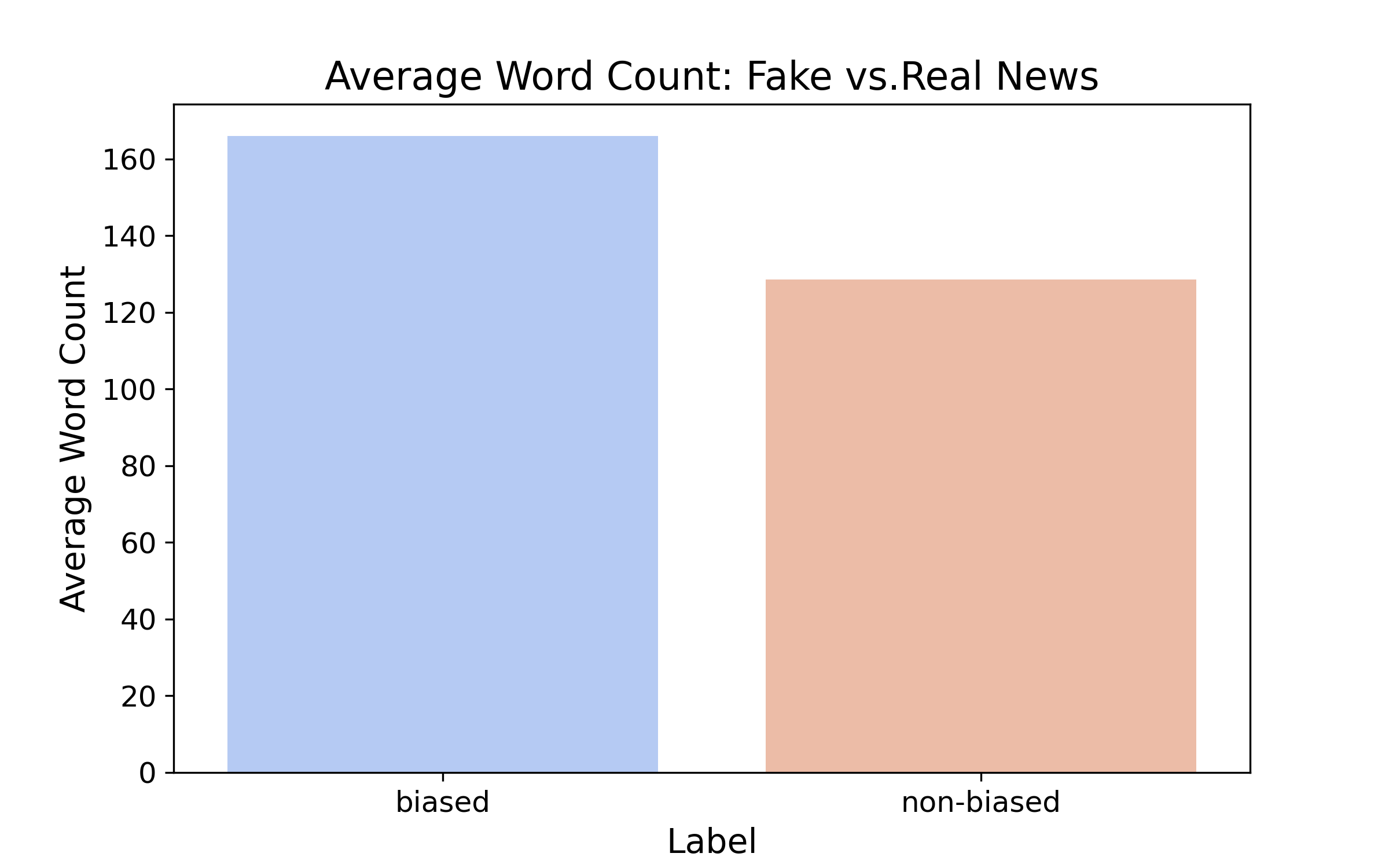}
        \captionsetup{width=1\linewidth}
        \caption{Average Word-Count:Fake vs. Real News}
        \label{fig:word_count_bias}
    \end{minipage}

\end{figure}

\begin{figure}[h]
    \centering
    \includegraphics[width=\linewidth]{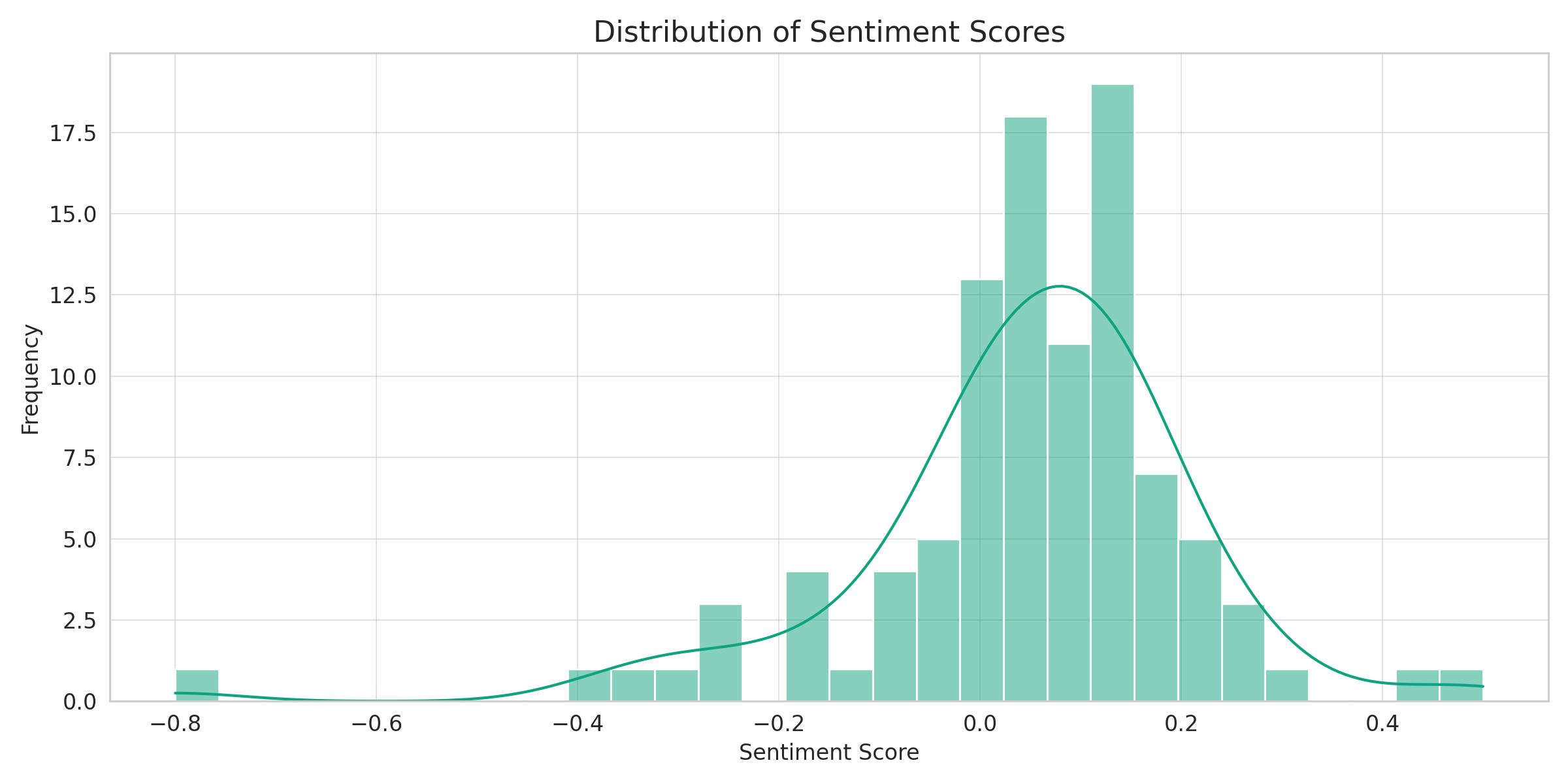}
    \captionsetup{width=0.8\linewidth}
    \caption{Distribution of Sentiment Scores in Fake News}
    \label{fig:sentiment_distribution}
\end{figure}

We show distribution of Fake vs. Real News in Figure \ref{fig:biased_non_biased}, and also see in Figure \ref{fig:word_count_bias} that average word count for fake news is around 160 words, while for real news it's approximately 140 words. 
  We also show the ``Distribution of Sentiment Scores'' (Figure \ref{fig:sentiment_distribution}) in the labelled, which is indicating a central tendency towards neutral sentiment with a slight positive skew. This may reflect a common trend in fake news \cite{allcott2017social}, where sentiment can be ambiguous, hence fake news is often subtle and tricky to discern.


\section{Benchmarking Classifier Performance}

We provide a comparison of various state-of-the-art classifiers for fake news detection (binary task). This benchmarking is performed to identify the most effective model in terms of various performance metrics such as Accuracy, Precision, Recall, F1-Score, and AUC-ROC. The classifiers included in this benchmarking are DistilBERT, ALBERT, BERT, and RoBERTa. Each model is configured with a Learning Rate of 0.0001, Batch Size of 16, and utilizes AdamW as the Optimizer. Each model is initially set to train for 20 epochs. However, we also employ early stopping to prevent overfitting and to optimize performance dynamically. All models share common hyperparameters including Warm-up Steps of 1000, a Max Sequence Length of 512, Weight Decay set at 0.01, Dropout Rate of 0.1, and Gradient Clipping at 1.0.  
 
\textit{Results:} The performance of different classifiers are shown in Table \ref{table:classifier_performance}. 
\begin{table}[ht]
\centering
\begin{tabular}{|l|c|c|c|c|c|}
\hline
\textbf{Classifier} & \textbf{Accuracy} & \textbf{Precision} & \textbf{Recall} & \textbf{F1-Score} & \textbf{AUC-ROC} \\
\hline
DistilBERT & 0.81 & 0.80 & 0.81 & 0.80 & 0.87 \\
ALBERT     & 0.80 & 0.78 & 0.85 & 0.81 & 0.87 \\
BERT       & 0.85 & 0.82 & 0.83 & 0.82 & 0.86 \\
RoBERTa    & 0.92 & 0.91 & 0.93 & 0.92 & 0.88 \\
\hline
\end{tabular}
\caption{Performance metrics of various classifiers}
\label{table:classifier_performance}
\end{table}

We observe that on average, RoBERTa outperforms the other models in all metrics. It shows particularly high scores in terms of Precision, Recall, and F1-Score. This suggests its high reliability and effectiveness in fake news detection tasks. BERT performs second-best in terms of overall performance that indicates its balanced performance in correctly classifying news items. ALBERT, while slightly comes behind BERT, shows decent Recall, which is an important factor for minimizing false negatives in news classification. DistilBERT, although the least performing in this comparison, still demonstrates adequate capability. It indicates that a distilled version can be used as an option where computational efficiency is a priority. Based on its best performance, we release the model weights of the RoBERTa fine-tuned model here \href{https://huggingface.co/newsmediabias/fake-news-classifier-elections}{newsmediabias/fake\_news\_elections\_classifier\_roberta} on Hugging Face.

\section{Discussion and Conclusion}
The fake news dataset presented in this study focuses on racial slurs and biases in North American political discourse. Advanced data extraction techniques were employed to include a wide range of news articles. These articles feature diverse keywords covering political, economic, social, cultural, and global themes. This variety ensures comprehensive coverage of crucial topics such as immigration reform, minorities rights, and environmental issues. Our data collection spanned from May to October 2023, capturing temporal variations. The collection comprised 40,000 articles, of which 4,000 were selected for detailed annotation. We utilized a combination of OpenAI's GPT-Turbo-3.5 API and expert human validation for this process. The dataset is structured with essential fields, including text, source, date published, and keyword categories.

Looking ahead, we plan to expand the dataset by adding more sources and extending the timeframe. This will ensure the dataset remains relevant, combat data and concept drifts and reflective of evolving discourse. Our future annotation efforts will integrate sophisticated AI models, such as newer LMs (e.g., LLamA2B or BARD), along with thorough human verification. 

Beyond its current use, the dataset offers valuable insights for academic research and practical applications. Its creation addresses the pressing need for comprehensive resources in studying misinformation and media biases. The topics it covers are critical in today's socio-political landscape, making the dataset an invaluable tool for researchers and practitioners. It aids in navigating and understanding the complexities of modern media discourse.

\section*{Data Availability}
The dataset used in this study is available for academic research. If you use this dataset, please cite this paper as a reference and contribute to the ongoing research efforts.

\bibliographystyle{plainnat}
\bibliography{references}

\end{document}